\pdfoutput=1
\documentclass[11pt]{article}

\usepackage{naacl2021}

\usepackage{times}
\usepackage{latexsym}

\usepackage[T1]{fontenc}
\usepackage[utf8]{inputenc}

\usepackage{microtype}

\newcommand{\method}{\textsc{RSL}\xspace}
\usepackage{microtype}
\usepackage{graphicx}
\usepackage{amsfonts,amssymb,amsmath}
\usepackage{mathtools}
\usepackage{tabu}
\usepackage{multirow}
\usepackage{algorithm}
\usepackage{algorithmic}
\usepackage{xspace}
\usepackage{booktabs} % To thicken table lines

\title{Reciprocal Supervised Learning Improves Neural Machine Translation}

\author{Minkai Xu\textsuperscript{*\rm 1,2}, Mingxuan Wang\textsuperscript{\rm 3}, Zhouhan Lin\textsuperscript{\rm 4}, Hao Zhou\textsuperscript{\rm 3}, Weinan Zhang\textsuperscript{\rm 4}, Lei Li\textsuperscript{\rm 3} \\
  \textsuperscript{\rm 1}University of Montreal \textsuperscript{\rm 2}Mila - Quebec AI Institute \textsuperscript{\rm 3}ByteDance AI Lab \textsuperscript{\rm 4}Shanghai Jiao Tong University\\
  \texttt{minkai.xu@umontreal.ca}\\
  \texttt{\{wangmingxuan.89,zhouhao.nlp,lileilab\}@bytedance.com}\\
  \texttt{lin.zhouhan@gmail.com}\\
  \texttt{wnzhang@sjtu.edu.cn}}

\begin{document}
\maketitle
\renewcommand{\thefootnote}{\fnsymbol{footnote}}
\footnotetext[1]{Work was done during the internship at Bytedance.}
\renewcommand{\thefootnote}{\arabic{footnote}}

\begin{abstract}

% Neural machine translation~(NMT) has achieved great success with the help of large amount of parallel data.
% However, different model architectures have different advantages and translation abilities, but it is hard to integrate them all together to one model.
% The ensemble method is too time-consuming for inference.
% Besides, monolingual data are also not fully utilized.
% Some works such as back-translation and unsupervised machine translation have tried to utilize monolingual data of target side, whereas the utilization of source side monolingual data still need be further explored.
% In this work, we propose a self-training framework with diverse teachers to make one model be able to learn advantages and diversities from other models, and monolingual data of source side language can also be utilized to further improve the translation performances.
% This method is very simple but much effective.
% Empirical results show that our method can obtain further improvements on the standard En$\to$De and En$\to$Fr translation tasks.

Despite the recent success on image classification, self-training has only achieved limited gains on structured prediction tasks such as neural machine translation (NMT). This is mainly due to the compositionality of the target space, where the far-away prediction hypotheses lead to the notorious \emph{reinforced mistake} problem. In this paper, we revisit the utilization of multiple diverse models and present a simple yet effective approach named \textbf{R}eciprocal-\textbf{S}upervised \textbf{L}earning (\method). \method first exploits individual models to generate \textit{pseudo parallel data}, and then cooperatively trains each model on the combined synthetic corpus. \method leverages the fact that different parameterized models have different inductive biases, and better predictions can be made by jointly exploiting the agreement among each other. Unlike the previous knowledge distillation methods built upon a much stronger teacher, \method is capable of boosting the accuracy of one model by introducing other comparable or even weaker models. \method can also be viewed as a more efficient alternative to ensemble. Extensive experiments demonstrate the superior performance of \method on several benchmarks with significant margins.\footnote{Code is available at \url{https://github.com/MinkaiXu/RSL-NMT}.}

\end{abstract}
\section{Introduction}
\label{sec:intro}

Recently, self-training method has shown remarkable success in image recognition. 
% Taking advantage of unlabeled data,
Trained on noisy augmented data, an EfficientNet model finetuned with self-training can achieve 87.4\% top-1 accuracy on ImageNet, which is 1.0\% better than the state-of-the-art model that requires 3.5B weakly labeled images~\cite{xie2019self}. Typically, in self-training we first train a base model on the labeled data, and then utilize the learned model to label unannotated data. Finally, both labeled and pseudo data are combined as the training set to yield the next level model. In the context of natural language processing, many works have successfully applied self-training technique including word sense disambiguation~\cite{yarowsky1995unsupervised} and parsing~\cite{mcclosky2006effective,reichart2007self,huang2009self}.

Nevertheless, the performance gains achieved through self-training are still limited for structured prediction tasks such as Neural Machine Translation~(NMT) where the target space is vast. Originally designed for classification problems, previous work suggests that self-training can be effective only when the predictions on unlabeled samples are good enough, and otherwise it will suffer from the notorious \textit{reinforced mistakes}~\cite{zhu2009introduction}. However, this problem is common in NMT scenario, where the hypotheses generated from a single model are often far away from the ground-truth target due to the compositionality of the target space~\cite{he2019revisiting}. \citet{zhang2016exploiting} found that training on this biased pseudo data may accumulate the mistakes at each time step and enlarge the error, and thus they propose to freeze the decoder parameters when training on the pseudo parallel data which may negatively impact the decoder model of NMT.

% We argue that the performance drop of self-training for NMT mainly comes from the \emph{reinforced mistakes}. 
To overcome this issue, in this paper we borrow the \textit{reciprocal teaching} concept~\cite{rosenshine1994reciprocal} from the educational field and revisit the core idea of classic ensemble approaches. Ensemble is built upon the assumption that different models have different inductive biases and better predictions can be made by majority voting. We propose to replace the self-supervision with \textbf{R}eciprocal-\textbf{S}upervision in NMT, leading to a novel co-EM (Expectation-Maximization) scheme~\cite{nigam2000analyzing} named \method. In \method, we use multiple separately learned models to provide diverse proper pseudo data, allowing us to enjoy the independence between different models and dramatically reduce the error through strategic aggregation.
%Most of these NMT works use only one type of neural network model such as ConvS2S~\citep{gehring2017cnnmt} and Transformer~\citep{vaswani2017transformer}.
%Usually, different neural models have different performances and they may also catch minor different patterns in the sequences.
More specifically, we first learn multiple different models on the parallel data. Then in the E-step all individual models are used to translate the monolingual data. And in the M-step the generated pseudo data produced by different models are combined to tune all student models.
%To combine these advantages and diversities, the intuitive method is ensemble, in which several models are trained and every model will be used during inference, then the output of these models are combined for a better prediction.

\method is inspired by the success of ensemble method. However, ensemble is resource-demanding during inference, which prevents its wide usage. Besides, it cannot make use of the large scale monolingual data from source side.
%The teacher-student framework can be used to make one model learn from others. Some works have been done to explore the assistance from right-to-left decoding model to usual left-to-right model~\citep{zhang2019regularizing,shan2019improving}. These works shown that a regular NMT model can learn from a right-to-left decoding model and obtain better performance. However, to our best knowledge, there are no such work exploring the assistance from several different models. So in this work, we try to utilize multiple different models as teachers, and train a student model to learn from them. Through this procedure, the student model can have better performance.
%Following this procedure, we have another advantage that monolingual data of source side language can be easily utilized to extend the training method to our self-training framework with diverse teachers.
%Similar to teacher-student framework for zero-shot NMT~\citep{chen-etal-acl2017-teacher}, the student model can also learn from teachers by monolingual data.
\method is also related to the data augmentation approaches for NMT. 
While most of previous works concentrate on monolingual data of target side such as back-translation~\citep{edunov2018understanding}, we pay more attention to the source side. Knowledge distillation (KD)~\cite{hinton2015distilling, mirzadeh2019improved} is another relevant research topic. However, KD is preliminary designed to improve a weak student model with a much stronger teacher model. By contrast, \method boosts the performance of base models through reciprocal-supervision from other just comparable or even weaker learners.
% Unsupervised machine translation~\citep{lample2018unsupervisedmt,artetxe2018unsupervisedmt,lample-2018-pbumt} can also be seen as utilizing target side monolingual data.
To the best of our knowledge, we are the first self-training framework with reciprocal-supervision, which can correct the bias of each model and fully utilize the monolingual data of source side language. More precisely, the advantages of \method
% our cooperative-supervised framework with diverse parameterized networks 
can be summarized as follows:
\begin{itemize}
    \vspace{-5pt}
    \item As an architecture-free framework, \method is flexible and promising, where a strong NMT model can benefit from any other comparable or even weaker models.
    \vspace{-6pt}
    \item \method is efficient for translation. While involving multiple models during training, only one NMT model is required for inference.
    \vspace{-6pt}
    \item Orthogonal to other widely used techniques such as back translation. \method is capable of making full use of the source side monolingual data.
    \vspace{-5pt}
    % which is a challenge problem.
\end{itemize}
Through extensive experiments, \method achieves significant gains on several standard translation tasks including En$\leftrightarrow$\{Ro, De\}. Surprisingly, we also have found that \method with other much weaker learners could even outperform a strong BERT enhanced NMT model with big margins.
\begin{figure}[!t]
    \centering
    \includegraphics[width=1.0\columnwidth]{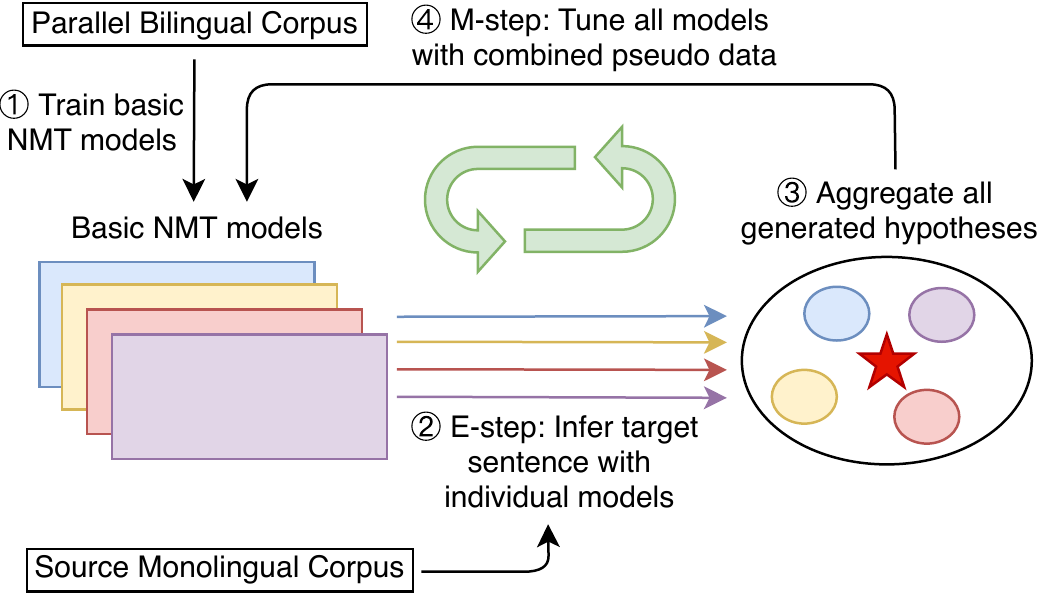}
    \caption{Illustration of the Reciprocal-Supervised NMT method. The red star represents the unknown gold target reference of the source side monolingual sentences. \method aggregate the hypotheses generated by multiple diverse models to serve as a better estimation of the pseudo data. 
    % \zhouhan{The green arrows' directions seem to be wrong.}
    }
    \label{fig:framework}
\end{figure}

\section{Proposed Method}
\label{sec:method}

% \minkai{add graphical illustration}

In this section, we first briefly introduce the background of modern NMT systems. Then we 
% elaborate the basic NMT models involved in our implementation of \method, and
present a detailed description of the proposed reciprocal-supervised learning framework. The whole framework is illustrated in Fig.~\ref{fig:framework}. Note that, \method is architecture-free, which can practically be adapted with arbitrary seq2seq models.

\subsection{Preliminary} 

Neural Machine Translation (NMT)~\cite{sutskever2014sequence,vaswani2017transformer} is a sequence-to-sequence (seq2seq)~\cite{bahdanau_iclr2018_nmt} learning framework to model the conditional probability $P(y|x)$ of the target translation $y$ given the source sentence $x$. 
% In practice, it is usually implemented with an encoder-decoder framework.
% , which can be parameterized as recurrent neural network~\cite{sutskever_nips2014_s2smt} and its variants~\cite{hochreiter1997long,cho2014learning,chung2014empirical}, convolutional neural network~\cite{gehring2017cnnmt,wu2018pay} or self-attention based transformer network~\cite{vaswani2017transformer,ott-etal-2018-scaling}, among which the recent self-attention based Transformer is the state-of-the-art architecture for NMT. 
% The encoder first map the source sentence into intermediate hidden vectors
% $h=(h_1,h_2,\dots,h_{T'})$ 
% in a continuous space, and then feed them into the decoder to generate target translation word-by-word. 
Given a parallel corpus $\mathcal{D}_{s,t}$ of source language $s$ and target language $t$, the training objective of the model is to minimize the negative likelihood of the training data:
\begin{equation}
% \footnotesize
    \mathcal{L}(\theta) = - \frac{1}{N} \sum_{(x,y)\sim \mathcal{D}_{s,t}}\log P(y|x;\theta),
\end{equation}
where $P(y|x;\theta)$ is the NMT model and $\theta$ denotes the model parameter. $N$ is the amount of parallel training data in $D_{s,t}$. 
Recently, there are plenty of works attempting to improve the performance of NMT system by new network architectures and modifications~\cite{luong2015effective,gu2016incorporating,hassan2018achieving}. Different from previous works, in this paper we take a step back and study how to make the most of multiple individual models by reciprocally learning.
% from each other. 
% Specifically, we investigate whether we can further boost the performance of a strong NMT model by introducing other comparable or even weaker networks.

\subsection{Basic NMT Models}
\label{subsec:basic}

\subsubsection{Diverse Parameterized Networks}

Modern NMT models can be parameterized with diverse different network architectures, such as recurrent neural network (RNMT)~\cite{sutskever2014sequence} and its variants~\cite{hochreiter1997long,cho2014learning,chung2014empirical}, convolutional neural network (Conv)~\cite{gehring2017cnnmt,wu2018pay} or self-attention based Transformer network~\cite{vaswani2017transformer,ott-etal-2018-scaling}, among which the recent self-attention based Transformer is the state-of-the-art architecture for NMT. In our \method framework, we hope different network architectures can capture different linguistic knowledge, and then all the knowledge can be aggregated into a single NMT model through the proposed \method approach. 
% In practice, we used \textbf{Transformer}~\citep{vaswani2017transformer,ott-etal-2018-scaling}, \textbf{DynamicConv}~\cite{wu2019pay} and \textbf{Hybrid model}~\cite{chen-etal-2018-rnmtplus} as the basic models.

\subsubsection{Bidirectional Decoding Factorization}
\label{subsubsec:r2l}

Seq2seq models are direction-sensitive.
Usually, the inputs of the encoder and the outputs of the decoder are both from left to right, leading to the \textit{exposure bias} problem~\cite{bengio2015scheduled}: during inference, true previous target tokens are unavailable and replaced by tokens predicted by the model itself, and thus mistakes made early can mislead subsequent translation, yielding unsatisfied translations with good prefixes but bad suffixes. To mitigate this issue in our inferred pseudo parallel data, we also involve right-to-left (R2L) decoding models~\citep{zhang2019regularizing,shan2019improving} together with common left-to-right (L2R) model.
% ad nd we canvise also reverse the directions.
% We use Transformer-R2L, DynamicConv-R2L or Hybird-RNMT+-R2L to denote the corresponding model where the inputs are from left to right but the output from right to left. 
Precisely, for R2L models the references will be reversed during training, and for inference procedure the outputs are also reversed  after decoding to recover the normal sequence.

\subsection{Reciprocal-Supervised NMT}
\label{subsec:method}

% Different basic models have different inductive biases with different architecture or factorization, so they may catch different sequence patterns and predict different proper hypotheses. Our \method framework aims to make the most of these diverse basic models and transfer the knowledge amongst each other.
% , thus yielding better performances.
% Furthermore, the reciprocal supervision enables extra monolingual data of source side to be integrated into this framework to further boost the accuracy. In the rest of this subsection, we will give the detailed elaboration of the reciprocal learning framework.
% This method is straightforward but effective.

Suppose we have a parallel corpus $\mathcal{D}_{s,t}$ of source language $s$ and target language $t$, and a monolingual corpus $\mathcal{D}_{s}$ of language $s$. In our pipeline, we will first use the bilingual dataset $\mathcal{D}_{s,t}$ to train multiple basic models $\mathcal{M}_1, \mathcal{M}_2, \cdots, \mathcal{M}_k$ by standard maximum likelihood training method, where $k$ denotes the number of models. In practice, our basic models involve both different network architectures and different decoding factorizations. After training, we further tune them through the reciprocal-supervision from each other. In this procedure, both bilingual and monolingual data can be utilized.
% Specifically, for the trained basic model $\mathcal{M}_i$, it can translate the sentences of language $s$ of $\mathcal{D}_{s,t}$ to language $t$, and the source sentences and the translated target sentences can be paired to a new dataset $\mathcal{D}_{s,t}^{p,i}$.
% Similarly, it also can translate the sentences in $\mathcal{D}_{s}$ to language $t$ and produce another dataset $\mathcal{D}_{s,t}^{m,i}$.
% Then we have real parallel dataset $\mathcal{D}_{s,t}$, and artificial parallel dataset $\mathcal{D}_{s,t}^{p,i}$, $\mathcal{D}_{s,t}^{m,i}$ from each basic model $\mathcal{M}_i$.
% We can mix these dataset in some proportion and shuffle them to train a new model for better results.
More specifically, the models are tuned with both the real parallel dataset $\mathcal{D}_{s,t}$ and the supervision from other different models.
% , and it also try to be consistent with those teacher models.
Thus, the training objective consists of the standard maximum likelihood (MLE) term on real parallel dataset and the reciprocal-supervision agreement (RS) term with other models.
We use Kullback-Leibler~(KL) divergence to measure the gap between different models. More precisely, the training loss for the $i^{th}$ basic model can be written as:
\begin{equation}
% \footnotesize
\small
    \begin{aligned}
        &\mathcal{L}_{MLE}(\theta_i)
        = -\frac{1}{N}\sum_{(x,y)\sim \mathcal{D}\!_{s,t}}\log P(y|x;\theta_i), \\
        &\mathcal{L}_{RS}(\theta_i)
        = \\
        & \quad \frac{1}{M}\sum_{x\sim \mathcal{D}_{s}'} \sum_{j=1\!,j\neq i}^k\!\!\Big( KL(P(y|x;\theta_j)||P(y|x;\theta_i)) \\
        & \quad + KL(P(y|x;\theta_i)||P(y|x;\theta_j)) \Big)\\
        & \mathcal{L}(\theta_i) = \mathcal{L}_{MLE}(\theta_i) + \mathcal{L}_{RS}(\theta_i)
    \end{aligned}
\end{equation}
% \begin{align}\footnotesize
%         % \scalebox{0.5}{
%         \mathcal{L}&_{MLE}(\theta_i)
%         = -\frac{1}{N}\sum_{(x,y)\sim \mathcal{D}\!_{s,t}}\log P(y|x;\theta_i), \\
%         \mathcal{L}&_{RS}(\theta_i)
%         = \nonumber\\
%         & \frac{1}{M}\sum_{x\sim \mathcal{D}_{s}'} \sum_{j=1\!,j\neq i}^k\!\! KL(P(y|x;\theta_j)||P(y|x;\theta_i)), \\
%         \mathcal{L}&(\theta_i) = \mathcal{L}_{MLE}(\theta_i) + \mathcal{L}_{CS}(\theta_i)
%         % }
% \end{align}
where $\theta_i$ denotes the parameters of the basic model $\mathcal{M}_i$.
% $\theta_j$ denotes the parameters of the student model that will be optimized.
$\mathcal{D}_{s}'$ is the combination of source side language sentences of both bilingual dataset $\mathcal{D}_{s,t}$ and monolingual dataset $\mathcal{D}_{s}$.
$N$ and $M$ are the sample numbers of parallel data $\mathcal{D}_{s,t}$ and combined monolingual data $\mathcal{D}_{s}'$ respectively in one training epoch. Note that, the proposed cooperative training framework is flexible, which can be adopted either with or without the monolingual corpus.

\begin{table*}[!t]
\small
\centering
\begin{tabular}{l|c|c|c|c|c}
\hline Dataset & WMT14 EN$\leftrightarrow$DE & WMT16 EN$\leftrightarrow$RO & IWSLT16 EN$\leftrightarrow$DE & WMT14 EN$\to$FR & WMT18 ZH$\to$EN \\
\hline Parallel & $4.50$M & $0.62$M & $0.20$M (TED) & $36$M &$22$M \\
Non-parallel & $5.00$M & $1.00$M & $0.20$M(NEWS) & $72$M &$10$M\\
\hline Dev/Test & newstest2013/14 & newstest2015/16 & tst13/14 & newstest2013/14& newstest2017/18\\
\hline
\end{tabular}
\caption{Statistics of datasets.}
\label{table:datasets}
\end{table*}

According to the definition of KL divergence that
$
KL(P||Q) = \sum_x P(x)\log\frac{P(x)}{Q(x)},
$
we can get the gradient of $\mathcal{L}(\theta_i)$ with respect to $\theta_i$. More specifically, since $P(y|x;\theta_j)$ is  irrelevant to parameter $\theta_i$, the partial derivative of $KL(P(y|x;\theta_j)||P(y|x;\theta_i))$ with respect to $\theta_i$ can be written as:
\begin{equation}
\label{eq:grad:kl1}
% \footnotesize
% \small
    \begin{split}
        & \frac{\partial KL(P(y|x;\theta_j)||P(y|x;\theta_i))}{\partial\theta_i}\\
        =& -\sum_{y\sim Y(x)} \frac{1}{\partial\theta_i} \partial \left[P(y|x;\theta_j) \log\frac{ P(y|x;\theta_i)}{P(y|x;\theta_j)}\right]\\
        =& -\sum_{y\sim Y(x)} P(y|x;\theta_j) \frac{\partial\log P(y|x;\theta_i)}{\partial\theta_i}\\
        =& - E_{y\sim P(y|x;\theta_j)} \frac{\partial\log P(y|x;\theta_i)}{\partial\theta_i},
    \end{split}
\end{equation}
where $Y(x)$ is the space of all possible target translations given the source sentence $x$. The term $\frac{\partial\log P(y|x;\theta_i)}{\partial\theta_i}$ are the standard gradients to maximize the log-likelihood within the seq2seq NMT system. And the expectation part $E_{y\sim P(y|x;\theta_j)}$ can be simply approximated by sampling from the $j^{th}$ NMT model. Therefore, minimizing this reciprocal-supervision term is equal to maximizing the log-likelihood on the pseudo sentence pairs generated from other basic model.

And similarly, the partial derivative of $KL(P(y|x;\theta_i)||P(y|x;\theta_j))$ with respect to $\theta_i$ is calculated as follows:
\begin{equation}
\label{eq:grad:kl2}
% \footnotesize
    \begin{split}
        & \frac{\partial KL(P(y|x;\theta_i)||P(y|x;\theta_j))}{\partial\theta_i}\\
        =& -\sum_{y\sim Y(x)} \frac{1}{\partial\theta_i} \partial \left[P(y|x;\theta_i) \log\frac{ P(y|x;\theta_j)}{P(y|x;\theta_i)}\right]\\
        =& - E_{y\sim P(y|x;\theta_i)}\log \frac{P(y|x;\theta_j)}{P(y|x;\theta_i)} \frac{\partial\log P(y|x;\theta_i)}{\partial\theta_i}.
    \end{split}
\end{equation}
Similarly, for the calculation of the expectation $E_{y\sim P(y|x;\theta_i)}$, we can also sample from the $i^{th}$ NMT model itself. However, one should note that there are vital differences between the above two terms (Eq.~\ref{eq:grad:kl1} and Eq.~\ref{eq:grad:kl2}): 1) Pseudo sentence pairs are generated from other models in the former term, but
from the $i^{th}$ model itself in the latter one; 2) The latter term, used the density ratio $\log \frac{P(y|x;\theta_j)}{P(y|x;\theta_i)}$ to re-weight the pseudo pairs generated from itself. Actually, the second KL term can be viewed a improved version of self-training, which employ the calculated weights to penalize the incorrect pseudo parallel data.

To sum up, the whole partial derivative of the objective function $\mathcal{L}(\theta_i)$ with respect to $\theta_i$ can be written as:
\begin{equation}
% \footnotesize
\small
    \begin{split}
        & \frac{\partial\mathcal{L}_{MLE}(\theta_i)}{\partial\theta_i} = -\frac{1}{N}\sum_{(x,y)\sim \mathcal{D}_{s,t}}\frac{\partial\log P(y|x;\theta_i)}{\partial\theta_i} \\
        &\frac{\partial\mathcal{L}_{RS}(\theta_i)}{\partial\theta_i} = \\ 
        &\quad -\frac{1}{M}\sum_{x\sim \mathcal{D}_s^{\prime}}\sum_{j=1,j\neq i}^k \Big(E_{y\sim P(y|x;\theta_j)} \frac{\partial\log P(y|x;\theta_i)}{\partial\theta_i}\\
        & \quad + E_{y\sim P(y|x;\theta_i)}\log \frac{P(y|x;\theta_j)}{P(y|x;\theta_i)} \frac{\partial\log P(y|x;\theta_i)}{\partial\theta_i}\Big)\\
        & \frac{\partial\mathcal{L}(\theta_i)}{\partial\theta_i} = \frac{\partial\mathcal{L}_{MLE}(\theta_i)}{\partial\theta_i} + \frac{\partial\mathcal{L}_{RS}(\theta_i)}{\partial\theta_i}.
    \end{split}
\end{equation}
In the final derivative $\frac{\partial\mathcal{L}_{MLE}(\theta_i)}{\partial\theta_i}$, the first MLE term can be easily optimized by standard maximum likelihood training on parallel data. For the second cooperative term, it can be approximately optimized by co-EM~\cite{nigam2000analyzing} algorithm,
% training each basic model on the sampled translated pseudo sentences from other basic models. This can be viewed as a EM algorithm, 
where we first estimate the expectation of target translation probability
$p(y|x)$ in the E-step, and then maximize the likelihood in the M-step. More specifically, in the E-step, we employ the multiple diverse NMT models to individually generate the target translations of source monolingual data, and then paired them to serve as the estimated new distribution of training data. In the following M-step, maximizing the likelihood is approximated by maximizing the log-likelihood of each individual model on the pseudo data. The above procedure will be repeated for several iterations until convergence, where the synthetic training data are re-produced with the updated NMT models.

% As shown in previous work, injecting noise is effective on both images~\cite{vincent2008extracting,xie2019self} and natural languages~\cite{edunov2018understanding,wu2019exploiting,he2019revisiting}. Therefore, in \method we also involve an additional noisy training part on the pseudo data. More specifically, we employ the following perturbation function $g(x)$: (1) random replace a word in the sentence to be a special unknown token ``$<\!\!\mathrm{UNK}\!\!>$" with probability 0.1; (2) randomly drop the words in all positions with probability 0.1; (3) randomly shuffle (swap) the words in the sentence with constraint that each words will not be shuffled further than three positions distance. Then the noisy term can be written as: 
% \begin{equation}\small
% \begin{split}
%     \mathcal{L}&_{noise}(\theta_i)
%         = \\
%         & \frac{1}{M}\sum_{x\sim \mathcal{D}_{s}'} \sum_{j=1\!,j\neq i}^k\!\! KL(P(y|x;\theta_j)||P(y|g(x);\theta_i))
% \end{split}
% \end{equation}
% And the total training loss for each involved model $\mathcal{M}_i$ can be written as $\mathcal{L}_{MLE}(\theta_i)+\mathcal{L}_{RS}(\theta_i)+\mathcal{L}_{noise}(\theta_i)$.

\begin{table*}[!t]
\small
\centering
% \resizebox{\linewidth}{!}{
\begin{tabular}{lcccccc}
% \hline \multirow{3}{*} { Model } & \multicolumn{2}{c|} { Low-RESOURCE } & \multicolumn{2}{c} { Rich-RESOURCE } \\
% \cline { 2 - 5 } & \multicolumn{2}{c|} { WMT16 EN $\leftrightarrow$ RO } & \multicolumn{2}{c} { WMT16 EN $\leftrightarrow$ DE }\\
% \hline 
\toprule
\multirow{2}{*} { Model } & \multicolumn{2}{c} 
{ WMT16 En $\leftrightarrow$ Ro } & \multicolumn{2}{c} { IWSLT16 En $\leftrightarrow$ De } & \multicolumn{2}{c} { WMT14 En $\leftrightarrow$ De }\\
% \cline { 2 - 7 }
& En-Ro & Ro-En & En-De & De-En & En-De & De-En \\
% \hline 
\midrule
Transformer & 32.1 & 33.2 & 27.5 & 32.8 & 28.6 & 31.3 \\
Transformer-R2L & 31.7 & 32.7 & 26.9 & 32.1 & 27.8 & 30.9 \\
% GNMT (Shah \& Barber, $2018)$ & 32.4 & 33.6 & 28.0 & 33.2 & 17.4 & 20.1 \\
% \hline GNMT-M-SsL + non-parallel (Shah \& Barber, 2018) & 34.1 & 35.3 & 28.4 & 33.7 & 22.0 & 24.9 \\
% \hline 
\midrule
Transformer+BT & 33.9 & 35.0 & 27.8 & 33.3 & 29.6 & 33.2 \\
Transformer+ST+BT & 34.2 & 35.4 & 28.2 & 33.7 & 30.4 & 34.2 \\
% Transformer+JBT + non-parallel~\cite{zhang2018joint} & 34.5 & 35.7 & 28.4 & 33.8 & 21.9 & 25.1 \\
% Transformer+Dual + non-parallel~\cite{he_nips2016_dualnmt} & 34.6 & 35.7 & 28.5 & 34.0 & 21.8 & 25.3 \\
% \hline 
\midrule
ML-NMT~\cite{bi2019multi} & 34.5 & 35.6 & 28.3 & 33.7 & 30.6 & 34.4 \\
\midrule
\method~(ours) & \textbf{35.0} & \textbf{36.2} & \textbf{28.7} & \textbf{34.1} & \textbf{31.1} & \textbf{35.0} \\
% Transformer+HS-JBT + non-parallel~(ours) & & & 30.4 & 34.5 \\
% \hline MGNMT & 32.7 & 33.9 & 28.2 & 33.6 & 17.6 & 20.2 \\
% MGNMT + non-parallel & $\mathbf{3 4 . 9}$ & $\mathbf{3 6 . 1}$ & 28.5 & 34.2 & $\mathbf{2 2 . 8}$ & $\mathbf{2 6 . 1}$ \\
% \hline
\bottomrule
\end{tabular}
% }
\caption{BLEU scores on both low-resource and rich-resource translation tasks. Transformer and Transformer-R2L are representative of basic models. +BT denotes using back translation to augment the training data, and +ST means further using the source side monolingual data through self-training. ML-NMT and \method represent instead using the multi-agent learning and proposed reciprocal learning to utilize the source side data respectively.}
\label{table:low-resource}
\end{table*}
\section{Experiments}
\label{sec:experiment}

\subsection{Experiment Setup}
\label{subsec:setup}
\noindent \textbf{Dataset.}
We compare our model with its counterparts in both resource-poor and resource-rich scenarios. For the former scenario, we conduct experiments on WMT16 English-to/from-Romanian (WMT16 En$\leftrightarrow$Ro)  low-resource task and IWSLT16 English-to/from-German (IWSLT16 En$\leftrightarrow$De) cross-domain task on TED talks. For the latter setting, we use WMT14 English-to/from-German (WMT14 En$\leftrightarrow$De) dataset. For all the settings, monolingual data are randomly picked from WMT News Crawl datasets. Furthermore, we also compare \method with some state-of-the-art large scale training methods on more widely used benchmarks, including WMT14 English-to-German and French (WMT14 En$\to$\{De,Fr\}) as well as WMT18 Chinese-to-English (WMT18 Zh$\to$En). The statistics about all datasets are listed in Tab.~\ref{table:datasets}. For data prepossessing, we use Moses scripts\footnote{\url{https://github.com/moses-smt/mosesdecoder}} for sentence tokenization, and \texttt{Jieba}\footnote{\url{https://github.com/fxsjy/jieba}} for Chinese sentence segmentation. Tokenized sentences of both source and target side are then segmented to sub-word units with Byte-Pair Encoding~(BPE)~\cite{sennrich-haddow-birch:2016:P16-12}\footnote{\url{https://github.com/rsennrich/subword-nmt}}. More specifically, we use a shared vocabulary of 32K sub-word units for En$\leftrightarrow$De and 40K units for En$\leftrightarrow$Fr, while use separated vocabularies of 32K tokens for En$\leftrightarrow$Ro and En$\leftrightarrow$Zh.

\noindent \textbf{Basic Models.} 
% \section{Basic Models}
In our implementation of \method, we employ several NMT models with different architecture to conduct the reciprocal learning process, including transformer, convolutional models and hybrid models. \textbf{Transformer}~\citep{vaswani2017transformer,ott-etal-2018-scaling} utilizes multi-head attention modules instead of recurrent units in encoder and decoder to summarize information of sentences. 
% More precisely, in transformer the attention weights are computed by comparing the current time-step vector to all elements in the context.
Besides, there are additional modules such as feed forward networks and layer normalization to consist stacked layers.
Transformer has shown its capacity on sequence modeling in lots of works.
% \noindent 
\textbf{Convolutional models} have been successfully applied on seq2seq tasks where the encoder and decoder are stacked convolutional layers without recurrent units.
ConvS2S~\citep{gehring2017cnnmt} is the first convolutional model with Gated Linear Unites~(GLU) and 
% learnable positional embeddings are applied, and every decoder layer has a 
separate attention blocks interacting with the outputs of the encoder.
% ConvS2S outperforms recurrent models before and achieves higher speed.
DynamicConv~\cite{wu2019pay} further improves the performance with sophisticated designed dynamic convolution kernels.
% \noindent 
\textbf{Hybrid model}~\cite{chen-etal-2018-rnmtplus} implement RNMT+ that combines the advantages of both recurrent networks and self-attention mechanism.
% This model uses LSTM~\citep{hochreiter1997long}, multi-head attention, and other modules to combine the advantages.
They further construct hybrid methods that use transformer or RNMT+ as encoder or decoder.
In \method, we use a Hybrid-RNMT+ model in which the encoder is transformer and the decoder is RNMT+, which shows the best performance among all hybrid models shown in their work.

\noindent \textbf{Model Settings.}
% In our implementation of \method, we use 
All our models (Transformer, DynamicConv and Hybrid-RNMT+) are implemented with fairseq\footnote{\url{https://github.com/pytorch/fairseq}}~\citep{ott-etal-2019-fairseq} in PyTorch~\citep{pytorch_nips}.
For the hyperparameters we mainly follow their original papers, and we also adopted some settings from \cite{ott-etal-2018-scaling}. More specifically, the dropout rate is set as 0.3 for all experiments, and all models are optimized with Adam~\cite{kingma2014adam} following the default optimizer settings and learning rate schedule in \cite{ott-etal-2018-scaling}.
All the models are trained on 8 NVIDIA V100 GPUs with half-precision~(FP16), and the batch size is set as 4096.
% For time efficiency, the newly trained models are fine-tuned from the best checkpoint of basic models, which do not hurt the experiment conclusions.
During inference, we generate translations with a beam size of 5 and a length penalty of 0.6.
For evaluation, we use sacreBLEU\footnote{\url{https://github.com/mjpost/sacreBLEU}}~\cite{post2018call} to report the case-sensitive detokenized BLEU score for WMT14 En$\leftrightarrow$De, WMT14 En$\to$Fr and WMT18 Zh$\leftrightarrow$Ee, and uncased tokenized BLEU for other tasks.
% WMT16 En$\leftrightarrow$Ro and IWSLT16 En$\leftrightarrow$De.

% For the sample number $N$ and $M$ introduced in Section~\ref{subsec:method}, we roughly choose $kM/N = 2$ from some validation experiments, and $k$ is the teacher number.
% That is, approximately twice training signals are from monolingual data.
% We sample from monolingual data without repetition in one epoch, but since the parallel data are limited, samples from parallel data may have repetitions.

% \subsection{Results}
\subsection{Numerical Results}
\label{sec:results}

We first present the overall performances of \method with its related baselines on several widely used datasets, and then make a comparison with several strong prior works:
% (1) \textbf{Basic models}: We choose Transformer and Transformer-R2L as representative of state-of-the-art bidirectional decoding model (see Section~\ref{subsubsec:r2l}) respectively, which shows the best BLEU score among all basic models in our experiments.
% (2) \textbf{Previous semi-supervised approaches (+BT and +ST)}: These baselines are trained on the combination of both bilingual parallel data and pseudo data generated from monolingual sentences by back translation (+BT) or forward translation, \textit{i.e.}, self-training (+ST).

\begin{itemize}
    \vspace{-4pt}
    \item Basic models: In practice, we find Transformer shows the best performance among all basic models in both bidirectional decoding direction, and thus we take it as the representative of state-of-the-art basic models.
    \vspace{-6pt}
    \item Previous semi-supervised approaches (+BT and +ST): These baselines are trained on the combination of both bilingual parallel data and pseudo data generated from monolingual sentences by back translation (+BT) or forward translation, \textit{i.e.}, self-training (+ST).
    \vspace{-6pt}
    \item Multi-agent Learning NMT (ML-NMT)~\citep{bi2019multi}: This recent proposed approach also introduces diverse agents in an interactive updating process, where each NMT agent learns advanced knowledge from an ensemble model during the training time.
    \vspace{-4pt}
\end{itemize}

The main results are shown in Tab.~\ref{table:low-resource}. In practice, we use the aforementioned 6 different models (3 different architectures $\times$ 2 different decoding directions as shown in Section~\ref{subsec:basic}) as basic NMT models to conduct reciprocal-supervised learning. We can see from the table that:
1) \textit{\method is robust in a range of different scenarios}. Sophisticated designed experiments show that \method is effective on several different tasks, including low-resource, rich-resource and cross-domain translations. These results suggest that as a generalized method, \method is not over-fitted to any specific dataset or translation task.
\begin{table}[!b]
\small
\centering
\begin{tabular}{lccc}
\hline
 SYSTEMS& En$\to$De & En$\to$Fr &Zh $\to$ En \\ 
 \hline
Transformer~\shortcite{vaswani2017transformer} & 28.4 & 41.0 & 24.1 \\
WMT Winner~\shortcite{wang-etal-2018-tencent}& - & - & 29.2 \\
BERT-NMT~\shortcite{yang2019towards} & 30.1 & 42.3 & - \\
MASS~\shortcite{song2019icml-mass} & 28.9 & -& - \\
XLM~\shortcite{lample2019cross} & 28.8 & -& - \\
mBERT~\shortcite{devlin2018bert} & 28.6 & -& - \\
ALM ~\shortcite{yangalternating} & 29.2 & -& - \\
\hline
\method & \textbf{31.1} & \textbf{43.4} & \textbf{29.4} \\
\hline
\end{tabular}
\caption{The performance comparison with other large-scale pretrained works on WMT 2014 En$\to$De, En$\to$Fr and WMT 2018 Zh $\to$ En. }
\label{table:comp}
\end{table}
2) \textit{\method is orthogonal to the widely adopted back translation technique}. Results show that \method works well with the classic BT technique, which promotes its wide usage.
3) \textit{\method can make the most of monolingual data}. While both self-training, multi-agent learning and reciprocal-supervised learning can further boost the performance of BT enhanced models, \method achieves much more noticeable gains. This meaningful observation roughly versifies the idea that diverse models are vital for higher quality pseudo data. This conjecture will be further discussed in Section~\ref{subsec:diverse-model}.

For training efficiency, we take IWSLT task as an example for illustration: with 1 Tesla V100 for each model, Transformer converges in \textasciitilde$14$ hours, Transformer+BT in \textasciitilde$21$ hours, Transformer+ST+BT in \textasciitilde$22$ hours, ML-NMT in \textasciitilde$34$ hours and RSL in \textasciitilde$28$ hours. In practice, we found that in RSL, it is the on-fly pseudo data sampling process that is time-consuming. 
% These details will be added in paper. 
One possible way to improve the efficiency may be to sample and save pseudo translations in advance for the training of next epoch.

Furthermore, we compare the performances of our \method with some important prior works on more challenging tasks. Previous work includes:
\begin{itemize}
\setlength{\itemsep}{0pt}
\setlength{\parsep}{0pt}
\setlength{\parskip}{0pt}
\vspace{-4pt}
    \item WMT Winner~\citep{wang-etal-2018-tencent} is the winner system of WMT18 Zh$\to$En task, implemented with rerank and ensemble techniques over $48$ models.
    \item BERT-NMT~\citep{yang2019towards} is a BERT-enhanced NMT model. It introduces BERT features with three delicately designed tricks, which helps to capture the knowledge of large scale source side monolingual data through the pre-trained language model.
    \item MASS~\citep{song2019icml-mass} is a seq2seq method pretrained with billions of monolingual data. 
    % We conduct experiments with the codes provided by the authors.
    \item XLM~\citep{lample2019cross} utilizes large scale cross-lingual data to pre-train the NMT model.
\vspace{-4pt}
\end{itemize}

The results are presented in Tab.~\ref{table:comp}. As shown in the table: \textit{\method is much more efficient in making use of the monolingual data}: while the pretrained language model is obtained through much larger monolingual or cross-lingual corpus, \method shows better results on these three benchmark translation tasks with much less cost.

\section{Analysis}

To further verify our conclusions, in this section we conduct several ablation studies. The majority of these empirical studies are conducted on WMT14 En$\to$De and En$\to$Fr datasets. 
% The En$\to$Fr dataset is constructed similar to the En$\leftrightarrow$De task, where the shared vocabulary size is 40K. The statistics of En$\to$Fr dataset are also incorporated in Tab.~\ref{table:datasets}.

% We conduct ablation studies on En$\to$De and En$\to$Fr translation tasks. We use the standard WMT 2014 dataset as the parallel training dataset.
% There are about 4.5 million sentence pairs for En$\to$De and about 36M sentence pairs for En$\to$Fr.
% The monolingual English data are randomly picked from all available WMT monolingual News Crawl datasets from year 2007 to 2017.
% We use newstest2013 as valid set and newstest2014 as test set for both En$\to$De and En$\to$Fr.

% \subsection{Results and Analyses}

\subsection{Are Diverse Models Necessary?}
\label{subsec:diverse-model}

This work is motivated by the conjecture of \textit{ensemble learning}, which assumes different models can capture different patterns of the sequence. Thus, the prediction mistakes of each model can be corrected by others, leading to a better estimation of the synthetic corpus constructed upon the source monolingual data. In this section, we implement \method with different basic models to investigate whether the diversity of basic models is necessary. The results are reported in \ref{table:model-arc}.

% Note that, in this comparison we concentrate on the utilization of source side monolingual data, so all experiments are conducted without back translation.

% We first study the Results presented in Tab.~\ref{table:model-arc}. Add a figure to show the bleu with each other to highlight the importance of diversity.

\begin{table}[!t]
% \small
\centering
% \resizebox{1\linewidth}{!}{
\begin{tabu}{lc}
\tabucline[0.65pt]{-}
\multicolumn{1}{c}{Model} & BLEU \\ \hline
Self-Training & 29.0 \\
\method w/ 6*Transformer & 29.3 \\
\method w/ Transformer(L2R+R2L) & 29.5 \\
\method w/ 3*Transformer(L2R+R2L) & 29.7 \\
\method w/ Heterogeneous Models & \textbf{30.1} \\
\tabucline[0.65pt]{-}
\end{tabu}
% }
\caption{The performance of \method on WMT 2014 En$\to$De with different basic models. Note that, in this comparison we concentrate on the utilization of source side monolingual data, so all experiments are conducted \textbf{without BT}. \textbf{L2R+R2L} denotes \method where basic models are Transformer and Transformer-R2L. \textbf{Heterogeneous Models} represents the standard implementation in Section~\ref{sec:experiment} where basic model are 6 different models. To yield a fair comparison we also incorporate the setting that basic models are 6 Transformers or 3 bidirectional Transformers, which are obtained by training with different seeds.}
\label{table:model-arc}
\end{table}

Empirical results suggest that though increasing the number of models (6*Transformer) can boost the performance, the improvement is rather limited compared with involving more diverse models (L2R+R2L and Heterogeneous settings). To investigate the vital element behind diverse models, we further measure the difference between different basic models through BLEU score. More specifically, we use each model to translate the test set into target space, and then compute the BLEU score between the predictions of different models. The BLEU scores between different architectures are about 56, and between different decoding direction are about 53. By contrast, the differences between the homogeneous models trained with different seeds are about 60, which is much higher. This observation indicates that heterogeneous models can produce more diverse hypotheses, which plays the vital role in our reciprocal learning framework.

\begin{table}[!b]
% \small
\centering
\begin{tabu}{lc}
\tabucline[0.65pt]{-}
\multicolumn{1}{c}{Pseudo Data} & BLEU \\ \hline
Beam Search (Top1) & 29.0 \\
Beam Search (Top3) & 28.6 \\
Diverse Beam Search (Top3) & 28.8 \\
\tabucline[0.65pt]{-}
\end{tabu}
\caption{The performance of \method with diverse generated pseudo data from single model. Topk means we generate k target predictions for each source side sentence. Beam Search and Diverse Beam Search~\cite{vijayakumar2016diverse} are two different decoding method, and the latter one is a diversity-promoting version of the former one.}
\label{table:diverse-sampling}
\end{table}

However, this conclusion further leads to another interesting question: since the diversity is important, whether we can induce the diversity by sampling multiple pseudo data from a single model to improve the performance of self-training? We use a single Transformer model to generate multiple hypotheses for each source side monolingual sentence during inference, and take them all together to construct the pseudo data. However, the results are less satisfactory as shown in Tab.~\ref{table:diverse-sampling}. We empirically found that many translations are of low quality when generating multiple predictions, which may hurt the performance of the NMT model during tuning on the generated pseudo data. These results suggest that involving the high training computation cost for multiple models is necessary to generate both diverse and good pseudo data, which is consistent with the well-known \textit{no free lunch theorem}~\cite{wolpert2012no}.

\begin{table}[!b]
% \small
\centering
% \caption{BLEU scores on resource-rich language pairs.}
\begin{tabular}{lcc}
\hline & \multicolumn{2}{c} {$\mathrm{En} \rightarrow \mathrm{De}$} \\
\cline { 2 - 3 } & newstest14 & newstest15 \\
\hline Transformer & 28.6 & 30.9 \\
Transformer w/ E & 29.5 & 31.7  \\
Heterogeneous w/ E & 29.7 & 32.1  \\
\method & 29.6 & 31.9 \\
\method w/ E & \textbf{30.2} & \textbf{32.5} \\
\hline
\end{tabular}
\caption{The performance of ensemble and \method. To yield a fair comparison with ensemble, all experiments are conducted without any source or target side monolingual data. Transformer w/ E means ensemble over 3 Transformers together, while Heterogeneous w/ E ensemble over the 3 different models in Section~\ref{subsec:setup}. \method w/ E means ensemble all the models after the reciprocal-supervised learning procedure.}
\label{tab:direct-ensemble}
\end{table}

\subsection{Direct Ensemble vs. \method}

% \paragraph{Ensemble Heterogeneous Models}
\method involves multiple models during reciprocal training, while uses just each individual model for inference. By contrast, in ensemble methods different models are separately trained, but are aggregated together during inference. Despite the high inference cost of ensemble method, it is interesting to compare the empirical performance of ensemble learning and the proposed \method. The results are shown in Tab.~\ref{tab:direct-ensemble}. As shown in the table, \method can achieve comparable performance against ensemble method with much less inference cost. Surprisingly, we found that ensemble over the reciprocal-supervised Learned models can further boost the performance. Our results suggest that: (1) The direct ensemble and \method can be viewed as a trade-off between training and testing cost. (2) \method is parallel to the ensemble method, and therefore better accuracy can be achieved by combining the two methods together.

% \method shows better performance than directly ensemble the different networks.

% \begin{table}[!ht]
% \centering
% \begin{tabu}{lcc}
% \tabucline[0.65pt]{-}
% \multicolumn{1}{c}{Model} & BLEU & time \\ \hline
% Transformer & 29.49 & \\
% Heterogeneous Models & 29.70 & \\
% \method & 30.01 & \\
% \tabucline[0.65pt]{-}
% \end{tabu}
% \caption{The performance of ensemble learning and our proposed method.}
% \end{table}

\subsection{Could \method still be effective with much weaker models?}
\label{subsec:weak-models}

In this section, we investigate in \method framework, whether a strong Transformer can still benefit from other much weaker NMT models. We first show the performances of our chosen weak models on WMT 2014 En$\to$De task in Tab.~\ref{table:basic}.

%  and En$\to$Fr translation 

\begin{table}[!t]
% \small
\centering
\begin{tabu}{lcc}
\tabucline[0.65pt]{-}
\multicolumn{1}{c}{Model} & En$\to$De & En$\to$Fr \\ \hline
ConvS2S & 26.2 & 40.8 \\
Transformer & 28.6 & 42.3 \\
Transformer-R2L & 27.8 & 41.8 \\
Hybrid-RNMT+ & 27.7 & 41.3 \\
 \tabucline[0.65pt]{-}
\end{tabu}
\caption{The performances of our re-implemented base models on WMT 2014 tasks.}
\label{table:basic}
\vspace{-10pt}
\end{table}

\begin{table}[!b]
% \small
\centering
\begin{tabu}{lcc}
\tabucline[0.75pt]{-}
  \multirow{2}{*}{Teachers} & \multicolumn{2}{c}{Student-Performance} \\
   & Single & Avg \\ \hline
ConvS2S & 28.8 & 28.9 \\
Transformer & 29.0 & 29.0 \\
Transformer-R2L & 29.2 & 29.3 \\
Hybrid-RNMT+ & 28.8 & 29.0 \\
% \tabucline[0.1pt on 2pt off 2pt]{-}
\hline
T + T-R2L & 29.6 & 29.7 \\
All Four Teachers & 29.8 & 30.1 \\
\tabucline[0.75pt]{-}
\end{tabu}
\caption{The influence on Transformer with different reciprocal learners on WMT 2014 En$\to$De.
The \emph{single} and \emph{avg} denote the scores evaluated on the checkpoint before averaging and averaged checkpoint respectively, which we show here for better comparison.
The BLEU score of basic Transformer is \textbf{28.6} for the averaged checkpoint.
T + T-R2L denotes using two models of Transformer and Transformer-R2L.}
\label{table:teacher}
\end{table}

% In Section~\ref{subsec:basic}, we introduce four basic models and they have different performances.
We conduct comprehensive experiments to study the capacity of \method with weak NMT models.
We first take each individual model as the teacher model and use the Transformer as the student model.
% , and use all the source side sentences in the parallel data as monolingual data.
The results are shown in Tab. ~\ref{table:teacher}.
We can find that R2L model can bring more obvious performance with better diversities.
For other L2R models, while a stronger teacher leads to a better student, it can still be improved with much weaker teachers.
Combining Transformer and Transformer-R2L in \method, we get better results than learning with a single heterogeneous model.
When jointly training all four models with reciprocal supervision, even though ConvS2S and Hybrid-RNMT+ are weaker, adding them also can bring significantly higher results.
These results suggest that  \method  is flexible, where strong learners can even benefit from other weaker models.

\subsection{Influence of Monolingual Data Size}

\begin{figure}[!t]
\centering
\includegraphics[width=0.85\columnwidth]{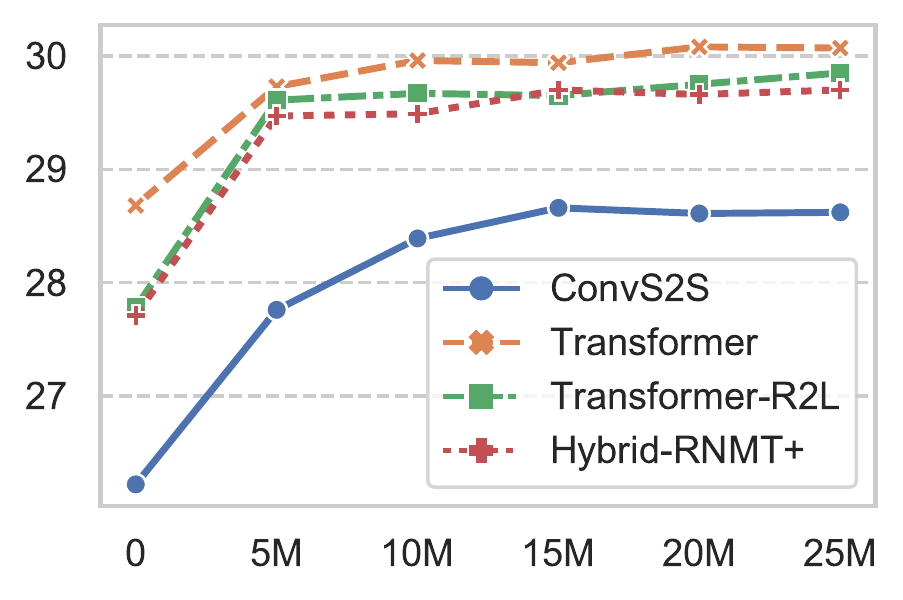}
\caption{The BLEU score influence with different monolingual data size on En$\to$De task. Four lines denote four different students.
% Each student is trained by all four teachers.
0 monolingual data means the performance of basic models.
}
\label{fig:num}
\end{figure}

In Figure~\ref{fig:num} we follow the En$\to$De task in Section~\ref{subsec:weak-models} and show how the number of monolingual sentences affects the performance of \method.
For different basic models, though have distinct performances, they receive a similar impact from monolingual data size. Results also show that monolingual data which are roughly two to three times as many as parallel data can already obtain satisfactory performances in this task.
% Our \method has great results compared with the baselines~(0 monolingual data).

% \subsection{Large Scale En$\to$Fr Results}

We study much larger scale monolingual data on En$\to$Fr task. 
% here we only use Transformer as the student model.
The results are shown in Table~\ref{table:enfr}.
We find that our \method can consistently achieve better performance with the larger monolingual corpus.

\begin{table}[!t]
% \small
\centering
\begin{tabu}{cc}
\tabucline[0.65pt]{-}
Monolingual data size & BLEU score \\ \hline
0 & 42.3 \\
36M & 43.0 \\
72M & 43.4 \\
 \tabucline[0.65pt]{-}
\end{tabu}
\caption{The performances of \method on WMT 2014 En$\to$Fr translation task with respect to monolingual data size. The sizes are multiples of parallel data.}
\label{table:enfr}
\vspace{-5pt}
\end{table}

% \subsection{Influence of Monolingual Data Split}
% In this section, we train \method with different monolingual data and report the results in ~\ref{table:zhen}. 

% \begin{table}[htb]
% \centering
% \begin{tabu}{lc}
% \tabucline[0.65pt]{-}
% Monolingual data size & BLEU score \\ \hline
% Transformer & 24.1 \\
% Transformer+BT+ST & 27.9 \\
% Transformer+BT+ST+ES*10 & 28.4 \\
% \method w/o DS & 28.2 \\
% \method w/ DS & 28.8 \\
% \method+ES & 29.4 \\
%  \tabucline[0.65pt]{-}
% \end{tabu}
% \caption{The performances of \method on WMT 2018 Zh$\to$En translation task with respect to monolingual data size. The sizes are multiples of parallel data size.}
% \label{table:zhen}
% \end{table}

% \subsection{Performance Comparison}

\section{Conclusion}
\label{sec:conclusion}

In this paper, we propose reciprocal supervised learning, an efficient and effective co-EM framework for neural machine translation. Different from previous methods, in \method a strong NMT model can benefit from any comparable or even weaker models, and the source monolingual corpus can also be fully utilized seamlessly.
Extensive experiments demonstrate the effectiveness and robustness of \method and provide insights on why and how \method can work well. 
\method is a general framework and can be extended for more NLP tasks, \textit{e.g.}, Q\&A, text summarization.
One potential direction for future work is to design better objective functions and set learnable weights for pseudo data from different models. Second, how to make \method more efficient is another interesting topic.
% By \method, the performances of neural network models are greatly improved and the experiments show the considerable results on standard En$\to$De and En$\to$Fr translation tasks.
% Our \method is rather simple but effective.

\newpage
\bibliography{ref}
% \bibliography{anthology,custom}
\bibliographystyle{acl_natbib}

\onecolumn
\newpage
\twocolumn
\appendix
\section{Related Work}

Our work is highly related to several important research directions of NMT.

% \subsection{Neural Machine Translation}
% \zhouhan{The NMT formulation should be moved out from related works section. The knowledge distillation part could be moved in here. }

% \subsection{Improving NMT with Monolingual Data}

\noindent \textbf{Improving NMT with Monolingual Data.} NMT heavily relies on a large amount of bilingual data with parallel sentence pairs, which is expensive to collect. To overcome this obstacle, many works have been proposed to leverage the rich monolingual data to help the training in the semi-supervised setting.
% , which can be unified into two directions. One idea is to 
\citet{gulcehre2015using,gulcehre2017integrating} incorporate external language models (trained separately on target monolingual data) into the NMT model, which improves the fluency in target language. \citet{sennrich-etal-acl2016-improvingwithmono} use the back translation (BT) approach to exploit target side monolingual data. They back translate the target side monolingual data to source side through an additional target-to-source NMT model learned on the bilingual dataset. Then 
% the generated data will be paired with monolingual data as 
the original bilingual data will be augmented with the synthetic parallel corpus for further training the source-to-target NMT model.
\cite{he_nips2016_dualnmt,xia2017dual,wang_iclr2018_multiagentnmt} learns from non-parallel data in a round-trip game via dual learning, where the source sentence is first forward translated to the target space and then back translated to the source space. The reconstruction loss is used to benefit the training.
\citet{zhang2018joint} propose to jointly train the source-to-target and target-to-source NMT models, where two models can provide back-translated pseudo data for each other.

While target side monolingual data has been extensively studied~\cite{poncelas2018investigating,cotterell2018explaining,edunov2018understanding,burlot2019using}, there exist few attempts to use the source side data. \citet{ueffing2006using} and \citet{zhang2016exploiting} explored self-training (ST) in statistical and neural machine translation respectively, albeit with limited gains. Recently, \citet{he2019revisiting} 
shows that the perturbation on the input and hidden states is critical for self-training on NMT. However, this study is conducted on relatively small-scale monolingual data, and therefore ST remains unclear in the large-scale setting.

\noindent \textbf{Ensemble and KD for NMT.} Among various model aggregating methods in machine learning, the most effective and widely adopted methods for NMT is the token-level ensemble.
% Let $\mathcal{X}$ and $\mathcal{Y}$ denote the source and target language spaces respectively, and $f_{m}: \mathcal{X} \mapsto \mathcal{Y}, m \in\{1, \ldots, M\}$ denotes the given $M$ source-to-target translation models. Let $\mathcal{V}_{t}$ denote the vocabulary of the target language.
In this approach, given the source sentences, a group of individually learned models cooperate together to generate the target sentence step by step. More specifically, the token-level ensemble method generates the target sequence by averaging the predicted probabilities of each token.

A commonly used KD approach is ensemble KD~\cite{fukuda2017efficient,freitag2017ensemble,liu2018distilling,zhu2018knowledge}, where each individual NMT model distill the knowledge from an ensemble model. More recently, multi-agent learning~\cite{bi2019multi} is proposed to distill the knowledge from a dynamic ensemble teacher model during training. By contrast, there is no teacher model and no ensemble procedure in \method, and we just improve individual models together through reciprocal learning. Note that, for \citet{bi2019multi}, models such as conventional Statistical MT or NMT with right2left decoding direction cannot be aggregated with common models with token-level ensemble method, which means \method is much more flexible.

% \subsection{Training Strategy}

\subsection{Discussions}

% Ensemble learning, which aggregates multiple diverse models during inference, has attracted huge interest in both academia and industry communities thanks to its effectiveness in a variety of computational intelligence problems such as classification~\cite{imamura2017ensemble}, prediction~\cite{imamura2017ensemble} and function approximation~\cite{fausser2011ensemble}. So far, many aggregating approaches have been developed such as bagging~\cite{breiman1996bagging} and (ada)boosting~\cite{kuznetsov2014multi} to improve the practical performance. 
% % Ensemble learning is primarily used to improve the classification task or reduce the likelihood of a poorly learned model. 
% Recently, ensemble of different neural networks \cite{hansen1990neural} has greatly improved the accuracy of neural machine translation (NMT), making it a vital widely used technique in state-of-the-art Neural NMT systems~\cite{sutskever2014sequence,bahdanau_iclr2018_nmt,vaswani2017transformer,hassan2018achieving}. In the scenario of NMT, a common implementation is to average the probability of each token computed by different individual models and then decode with the averaged probabilities. Previous studies show that the performance of ensemble method heavily depends on both the accuracy and diversity of base models~\cite{imamura2017ensemble,wangtransductive}, which are typically obtained through independent training on different sets of attributes.
% Clarify the relationship between \method and KD \& ensemble.

There are several essential differences between \method and other widely used methods that involves multiple models.

\noindent \textbf{Ensemble.} In the scenario of NMT, a common implementation of ensemble is to average the probability of each token computed by different individual models and then decode with the averaged probabilities. Despite its success in state-of-the-art Neural NMT systems~\cite{sutskever2014sequence,bahdanau_iclr2018_nmt,vaswani2017transformer,hassan2018achieving}, in practice there are a few common challenges of ensemble methods, which prevent its wide usage: 1) \textit{High computational cost}. For ensemble learning, all individual models have to conduct encoding and decoding, which is prohibitively time and memory consuming. This will be even worse in the context of NMT due to the large size of state-of-the-art networks like transformer~\cite{kitaev2020reformer}. 2) \textit{Absence of monolingual data.} Ensemble cannot make use of the large scale monolingual data from the source side. By contrast, \method involves the co-EM procedure in reciprocal-supervised learning, which can enjoy the benefit of monolingual data.

\noindent \textbf{Knowledge distillation (KD).} KD~\cite{hinton2015distilling} is another related topic, where we usually first train a strong teacher model and use it to give \textit{soft label} of the training data. Then a smaller model is trained with the \textit{soft label} to distill the knowledge from the strong model. For NMT, when multiple models are provided, a typical practice is to take the ensemble of all the models as the teacher model to translate the source side data into single pseudo data. By contrast, in \method we just use individual models to translate the monolingual data into multiple proper target sentences, which encourage the basic models to learn more diverse patterns. Furthermore, compared with KD, the absence of the \textit{teacher} role in \method allows the strong models to benefit from other comparable or even much weaker models.

\end{document}